\documentclass[runningheads]{llncs}

\usepackage{eccv}

\usepackage{eccvabbrv}
\usepackage{amssymb}
\usepackage{graphicx}
\usepackage{booktabs}
\usepackage{tabularx,multirow}

\usepackage[accsupp]{axessibility}  % Improves PDF readability for those with disabilities.

\usepackage[pagebackref,breaklinks,colorlinks]{hyperref}
\usepackage{orcidlink}

\begin{document}

\title{NeRO: Neural Road Surface Reconstruction} 

\titlerunning{NeRO: Neural Road Surface Reconstruction}

\author{Ruibo Wang\inst{1} \and
Song Zhang\inst{1} \and
Ping Huang\inst{2} \and
Donghai Zhang\inst{2}\and
Haoyu Chen\inst{2}} 

\authorrunning{R.Wang et al.}

\institute{
% Z-one Technology Co., LTD. \and
% Springer Heidelberg, Tiergartenstr.~17, 69121 Heidelberg, Germany
Z-one Technology Co., LTD. Shanghai, China\\
\email{\{wangruibo01,zhangsong05,huangping01,zhangdonghai,chenhaoyu03\}@saicmotor.com}}

\maketitle

\begin{abstract}
Accurately reconstructing road surfaces is pivotal for various applications especially in autonomous driving. This paper introduces a position encoding Multi-Layer Perceptrons (MLPs) framework to reconstruct road surfaces, with input as world coordinates x and y, and output as height, color, and semantic information. The effectiveness of this method is demonstrated through its compatibility with a variety of road height sources like vehicle camera poses, LiDAR point clouds, and SFM point clouds, robust to the semantic noise of images like sparse labels and noise semantic prediction, and fast training speed, which indicates a promising application for rendering road surfaces with semantics, particularly in applications demanding visualization of road surface, 4D labeling, and semantic groupings.
\end{abstract}

\section{Introduction}
\label{sec:intro}

The evolving landscape of 3D reconstruction has led to significant advancements, especially in reconstructing complex urban environments like road surfaces, which is useful in 4D labeling and semantic groupings. While effective, traditional approaches often grapple with challenges such as computational intensity, low-quality rendering, and semantic noise to be improved. 

NeRF presents a method for synthesizing novel views of complex scenes by modeling the volumetric scene function using a fully connected deep neural network. While NeRF's methodology provides high-quality 3D reconstructions, its voxel representations are redundant for surface modeling and computationally intensive. Tesla claimed on its 2021 AI Day that it uses implicit MLPs to reconstruct road surface color, semantics, and elevation without using any explicit meshes or point clouds but offering no further information. RoME \cite{mei2023rome} uses explicit mesh objects from Pytorch3D \cite{ravi2020pytorch3d} lib to represent the color and semantics and uses the implicit MLPs network with position encoding only to calculate the road height. After learning road height with supervised road ground truth like SFM point clouds or lidar point clouds, it utilizes ray-tracing based method in Pytorhc3D lib to optimize the color and semantics, accelerating the render speed than NeRF. 

\begin{figure}[tb]
  \centering
  \includegraphics[scale=0.5]{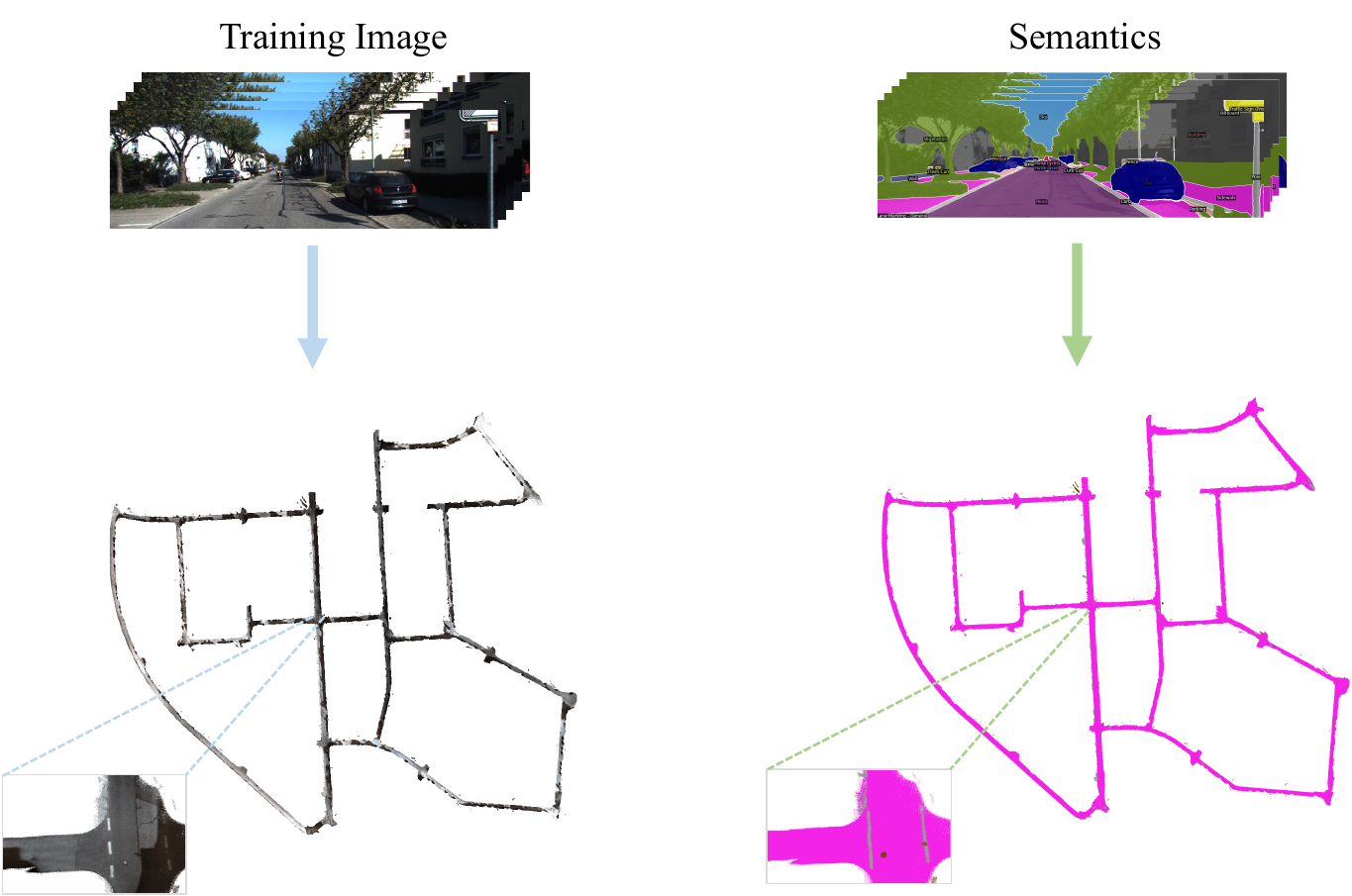}
  \caption{The result of our method in reconstructing an entire segment of the road from the KITTI Odometry Sequence 00. 
  }
  \label{fig:intro}
\end{figure}

We use a unified position encoding MLPs-based network architecture Neural Road Surface Reconstruction (\textbf{NeRO}) to reconstruct height, color, and semantics in a function like (height, color, semantics) = MLP(x,y). The ground truth of height could be derived from various sources like vehicle camera pose, LiDAR point clouds, and SFM point clouds.  The ground truth of color and semantics are queried by sampling millions of world coordinates road surface 3D points around each camera and projecting them back to related image coordinates using camera extrinsic and intrinsic, without utilizing ray-tracing based method. The main contributions of this paper include:
\begin{itemize}
\item We introduce a position encoding MLPs-based road surface reconstruction method to represent road surfaces with color, semantics, and height. Our method uses the same software libraries as NeRF without any further complex software, and it may be the closest implementation to Tesla's implicit road surface reconstruction.

\item We have verified the performance of our architecture in road height reconstruction under different input sources, such as vehicle camera pose, SFM point clouds, and LiDAR point clouds. We have also tested its capability to reconstruct slopes (using square waves as a substitute) and its ability to complete sparse point cloud inputs or incomplete height information. 

\item Additionally, we validated our architecture's robustness against semantic noise. Experiments show that semantic noise in single-frame images can be optimized to some extent by aggregating multi-frame, multi-view semantic information, which helps improve the accuracy of 4D road surface labeling.
\end{itemize} 
Our code will be released at: \url{https://github.com/ToeleoT/NeRO}

\section{Related Works}
\subsection{Multi-view 2D Image to 3D Reconstruction} 
A notable contribution in this realm is the rendering technique outlined in \cite{schonberger2016structure,schonberger2016pixelwise}. Those methods demonstrate a robust capability to reconstruct three-dimensional geometries from two-dimensional images, leveraging known camera poses. In the context of urban mapping, \cite{qin2021light} presents a technique that focuses on extracting road line segmentation data from images, followed by an inverse projection of this information to generate three-dimensional maps. A similar method \cite{waechter2014let} introduces a method for creating textured meshes from images. \cite{leroy2018shape} explores depth map creation from various viewpoints, which fuse them into a cohesive three-dimensional structure. \cite{sun2021phi,xu2020planar} capitalizes on the advantages of planar structures, thereby refining the efficiency and accuracy of three-dimensional reconstructions in certain contexts.
\subsection{NeRF-based Reconstruction}
Neural Radiance Fields (NeRF) \cite{mildenhall2021nerf} represents an implicit approach, utilizing MLPs and positional encoding method divergent by inputting viewing angles to reconstruct three-dimensional environments. Subsequent research has expanded upon the foundational principles of NeRF, adapting it for extensive, unbounded scenes. Specifically, NeRF++ \cite{zhang2020nerf++} employs a methodology of normalizing the unbounded scene into a unit sphere to render the environments. Further advancements are seen in BlockNeRF \cite{tancik2022block}, which deconstructs large-scale scenes into smaller blocks with each NeRF framework, incorporating factors of exposure and visibility to reconstruct these extensive environments. Conversely, Mip-NeRF 360 \cite{barron2022mip} introduces an innovative wrapping technique for parameterizing Gaussians. \cite{kundu2022panoptic} renders the background with a single MLP, and other dynamic objects are learned by a set of MLPs to present a panoptic street view. \cite{deng2022depth,rematas2022urban,xu2022point,li2022read} uses the depth map, LiDAR, and point cloud data to help NeRF to build more accurate 3D geometry.
\subsection{NeRF with Semantic}
There has been much work integrating semantic information into 3D scene reconstruction in the evolving landscape of the implicit method. A pivotal development in this domain is Semantic-NeRF \cite{zhi2021place}, which incorporates an additional layer within the MLPs of NeRF to render the semantic details within three-dimensional environments. Expanding upon this concept, PanopticNeRF \cite{fu2022panoptic} merges 3D semantic data with three-dimensional bounding boxes to enhance the geometric fidelity of semantic representations. Another innovative approach is presented in NeRF-SOS \cite{fan2022nerf}, which leverages self-supervised learning techniques within the NeRF framework to calculate semantic information. Furthermore, NeSF (Neural Semantic Fields) \cite{vora2021nesf} demonstrates a unique approach by extracting density information from NeRF outputs, which is processed through a 3D U-Net architecture.
\subsection{NeRF-based Road surface Reconstruction}
RoME \cite{mei2023rome} significantly contributes to the reconstruction of road surfaces by using explicit mesh objects from Pytorch3D \cite{ravi2020pytorch3d} lib to represent the color and semantics and using the implicit MLPs network with position encoding to calculate the road
height. After learning road height with supervised road ground truth like SFM and lidar, it utilizes ray-tracing based method in Pytorhc3D lib to go on optimizing the color and semantics.  MV-Map\cite{xie2023mv} adopts a voxel-based approach within the NeRF framework like \cite{yu2021plenoctrees,fridovich2022plenoxels}, paving the way for the construction of high-definition maps. Moreover, PlaNeRF \cite{wang2023planerf} presents a plane regulation methodology grounded in the decomposition (SVD) technique to reconstruct scenes efficiently. StreetSurf \cite{guo2023streetsurf} distinguishes itself by segmenting scenes within images and applying varied scales of multi-resolution hash positional encoding. 

\section{Method}
\begin{figure}[tb]
  \centering
  \includegraphics[scale=0.23]{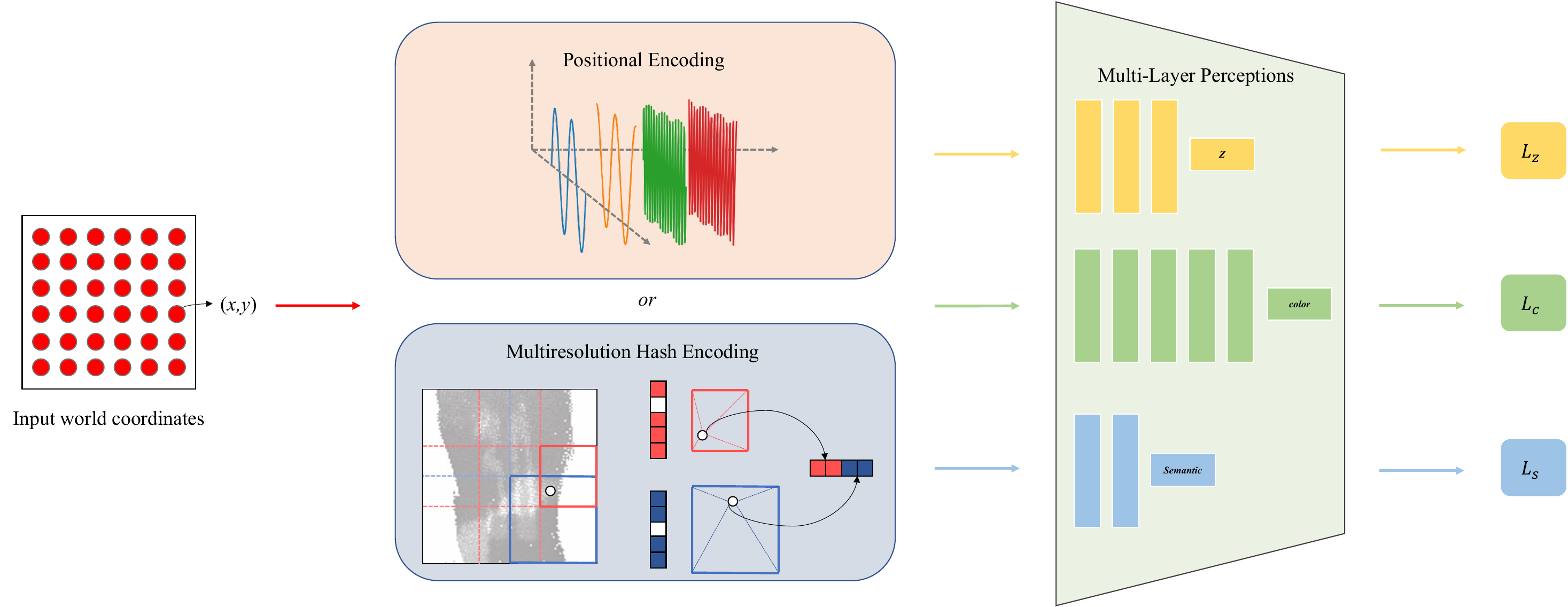}
  \caption{\textbf{NeRO Overview.} The world coordinates \textbf{X} = (\textit{x}, \textit{y}) are encoded by the Positional Encoding or Multiresolution Hash Positional Encoding. Then, the encoded information is passed into three different MLPs responsible for calculating the height z, color, and semantic.
  }
  \label{fig:method}
\end{figure}

We show our framework and pipeline in \cref{fig:method}. \textbf{NeRO} processes the input to calculate the height along the vertical z-axis, RGB, and semantic.
\subsection{NeRO Network Structure}
\label{sec:net}
\textbf{NeRO} takes the x and y coordinates in the world coordinate system, \textbf{X} = (\textit{x}, \textit{y}), as input. Before entering the network layers, our input \textbf{X} is normalized to between [-1,1] to facilitate the calculation of encoding methods. We feed the normalized input $\mathbf{X'}$ into position encoding functions. Then, the outputs from the position encoding methods are processed by three different MLPs, resulting in outputs for road surface height \textbf{z}, colour output \textbf{c} = (\textit{r}, \textit{g}, \textit{b}), and semantic output \textbf{s}. NeRO can also choose different position encoding method for different MLPs.  By the way, only MLPs without any position encoding fails to reconstruct road height, color and semantic through all of our experiments.

\subsection{Encoding methods}
\label{sec:enc}
\subsubsection{Positional Encoding}
From \cite{mildenhall2021nerf,muller2022instant,tancik2020fourier}, we need to leverage Positional Encoding, which is used to learn high-frequency details with the formula: 
\begin{align}
PE(X) = (sin(2^{0}\pi X), cos(2^{0}\pi X),...,sin(2^{L-1}\pi X), cos(2^{L-1}\pi X))
\end{align}
Where $L$ is the hyperparameter that decides the length of the function.

\subsubsection{Multi-Resolution Hash Positional Encoding}
In \cite{muller2022instant}, a learnable encoding method, Multi-Resolution 3D Hash Positional Encoding is introduced. Our approach modify a Multi-Resolution 2D hash Positional Encoding based on it.

\subsection{Reconstruction methods}
\label{sec:rec}
\subsubsection{Z-axis Reconstruction}
We employ ground truth height from three sources: vehicle camera pose, LiDAR point clouds, and SfM point clouds. In vehicle camera pose, it is assumed that the ground plane near the corresponding pose is flat, and each pose will sample the points within a length * width certain area to form pseudo point clouds. The different representations in the encoding method will affect the result of the height value.
\subsubsection{Color Reconstruction}
% 把loss部分提过来
In color reconstruction, we sample millions of 2D world coordinates as network input \textbf{X} = (\textit{x}, \textit{y}) for each pose. Then, we use those coordinates to obtain height $\mathbf{z}$ and color \textbf{c} by the complete learned height network and the color network separately. After that, we combine the road surface height $\mathbf{z}$ with \textbf{X} = (\textit{x}, \textit{y}) to obtain the 3D world coordinates \textbf{W}  = (\textit{x}, \textit{y}, \textit{z}), then project them into the pixel coordinate system \textbf{(u, v)} with camera's extrinsic and intrinsic to get the corresponding ground truth pixel color $\mathbf{c'}$ to optimize network output color $\mathbf{c}$.
\subsubsection{Semantic Reconstruction}
% 要说和color一样的步骤
The network semantic output $\mathbf{s}$ from the encoding methods is used to render the semantic information for the network input \textbf{X} = (\textit{x}, \textit{y}), which use the same method in color reconstruction to obtain the ground truth semantic $\mathbf{s'}$.

\subsection{Loss Function}
\label{sec:opt}

% \subsubsection{Z-axis Loss}
We calculate the Mean Squared Error loss for the height MLPs output \textbf{z} with the ground truth $\mathbf{z'}$, which is the $\mathcal{L}_{z}$. 
% \subsubsection{Color Loss}
Then, we use the true colour $\mathbf{c'}$ obtained from the pixel coordinate system and compare it with the network output colour $\mathbf{c}$, After that, it will perform Mean Squared Error loss calculation $\mathcal{L}_{c}$. 
% \subsubsection{Semantic Loss}
Finally, we use the semantic ground truth $\mathbf{s'}$ and the semantic output \textbf{s} to perform a Cross-entropy loss calculation $\mathcal{L}_{s}$.  

% \subsubsection{Overall Loss}
The total loss for our method is that:
\begin{align}
\mathcal{L} &= \mathcal{L}_z + \mathcal{L}_c + \mathcal{L}_s \\
& = \sum_{n = 1}^{N} \left[\Vert z - z'\Vert^{2}_{2} + \Vert c - c'\Vert^{2}_{2} + \sum_{l = 1}^{L} slog(s')\right]
\end{align}
Where $N$ is the number of points inside the road surface that appear in the road part of the training images, and $L$ is the number of semantic labels used in the training process.

\section{Experiment}
\label{sec:exp}
\subsection{Experiment Settings}
\subsubsection{Datasets}

\begin{figure}[tb]
{\centering%
\begin{tabular}{@{}c@{}}
\includegraphics[width=50mm]{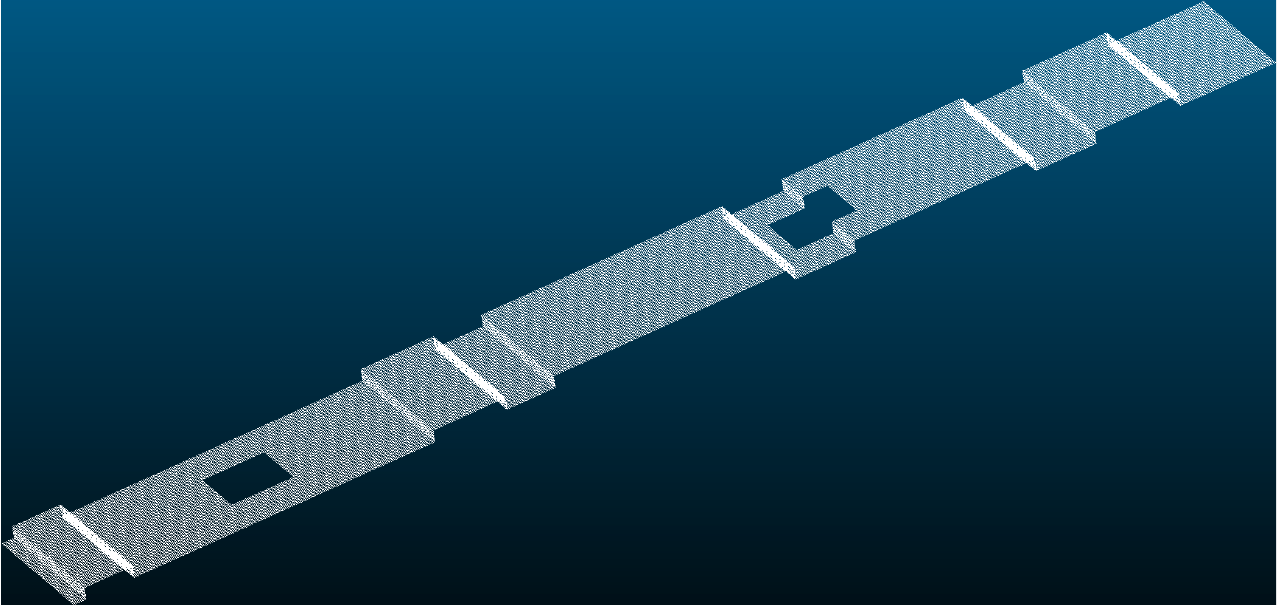}
\\
\tiny Ground Truth
\end{tabular}\par}
{\centering%
\begin{tabular}{@{}cc@{}}
\includegraphics[width=50mm]{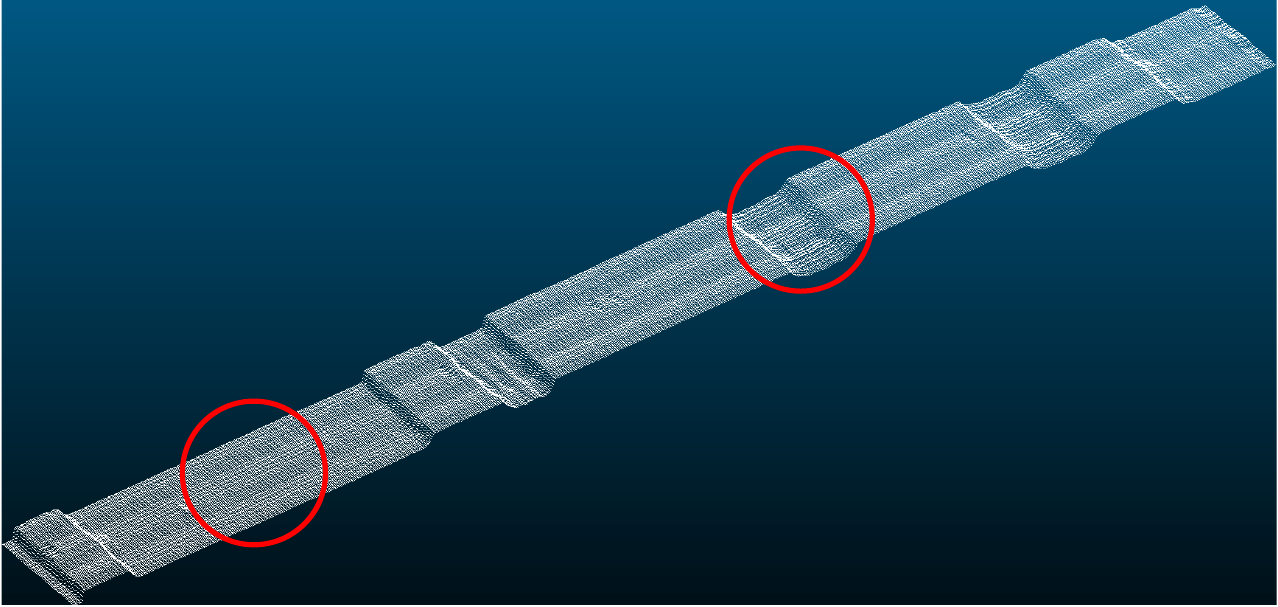}&
\includegraphics[width=50mm]{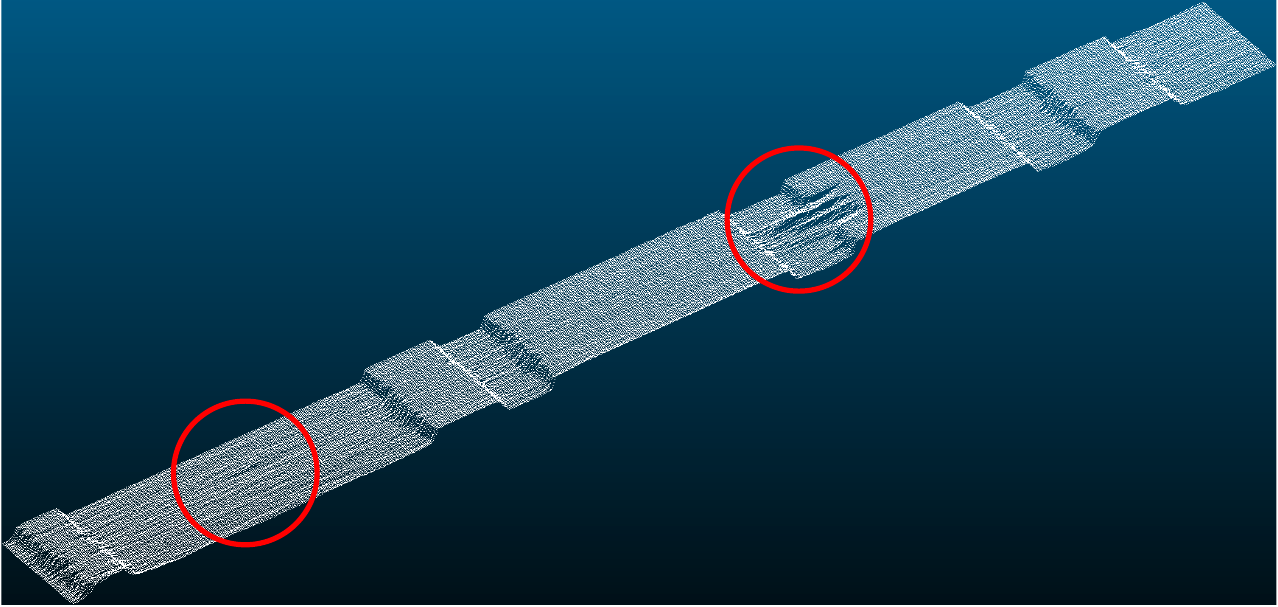}
\\
\tiny PE  & \tiny Hash PE
\end{tabular}\par}

{\centering%
\begin{tabular}{@{}cc@{}}
\includegraphics[width=50mm]{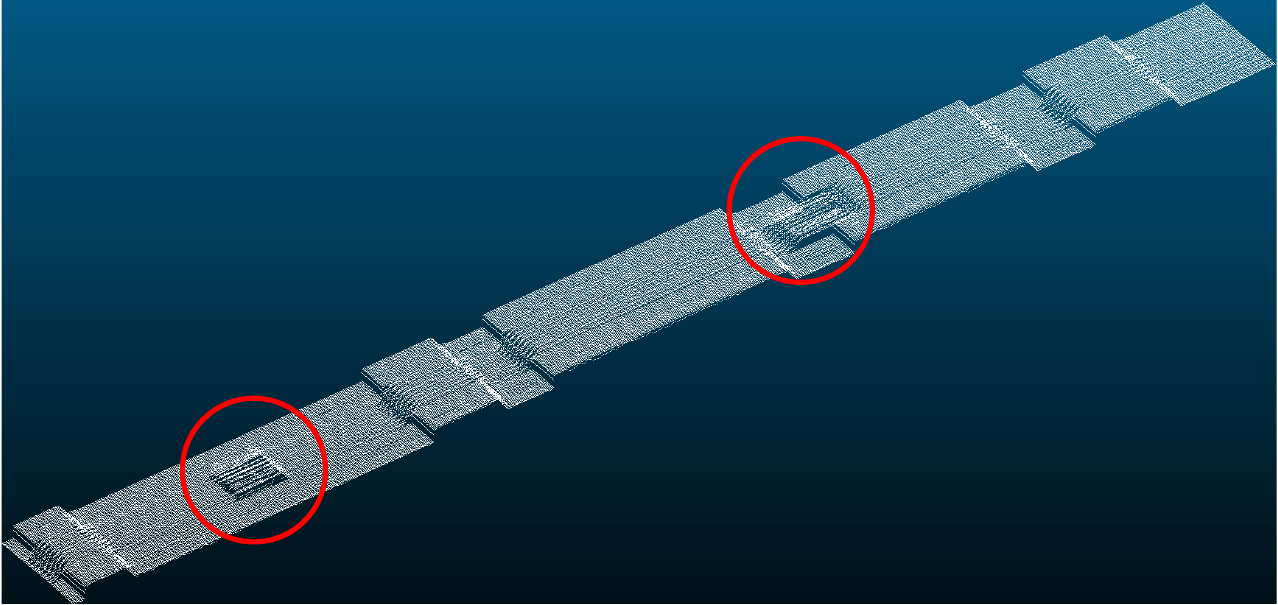}&
\includegraphics[width=50mm]{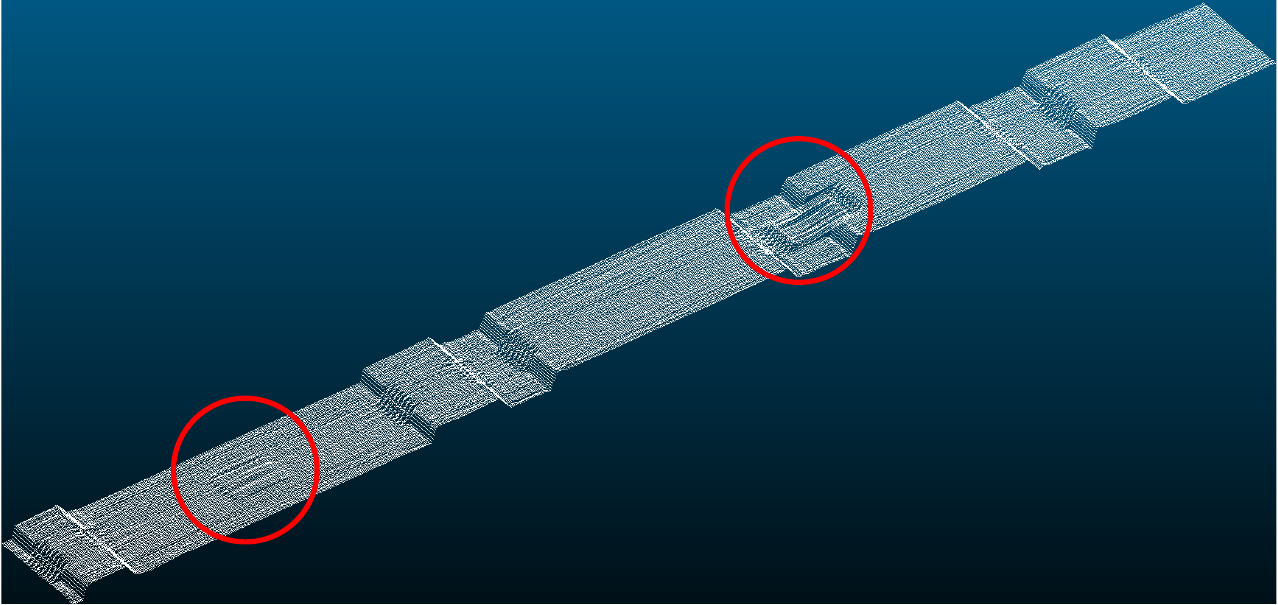}
\\
\tiny Hash PE single grid  & \tiny Muti-grid PE no Hash
\end{tabular}
\caption{Comparison in reconstructing the incomplete road. The red circle indicates the hole in the horizontal and corner of the road surface.}
\label{fig:hole}\par}
\end{figure}

Our experiment uses Kitti odometry \cite{geiger2012we} sequence 00 datasets to test our method, and we use images from the left camera as our training datasets. As the Kitti odometry datasets do not have any semantic information, we use the state-of-the-art semantic prediction method Mask2Frormer \cite{cheng2022masked} with Swin-L \cite{liu2021swin} backbone to acquire that information. The semantic categories we used are road, traffic lane and Manhole, which represent the main information appears in road. The SfM points are obtained by COLMAP, which uses the provided but not perfect ground truth camera pose in kitti odometry datasets for reconstruction.
\subsubsection{Evaluation Metrics}
Our experiment compares the pixel color from the ground images, and use PSNR as metric to verify the rendering quality. For semantic label accuracy, the metric is mIoU.
\subsubsection{Environment Setup}
We implements NeRO method by Pytorch \cite{paszke2019pytorch} framework, which uses the Nvidia A100 80G for the experiments, while a GPU with 8G memory is possible. The training images are the original fixed size, 1241 x 376, with batch size 1. The training optimizer we selected is Adam \cite{kingma2014adam}, with a learning rate of 5e-4.

\subsection{Road Surfaces Height Reconstruction with different PE Method} To illustrate the road height reconstruction performance, an experimental setup was set utilizing a square wave-like road structure, as depicted in the \cref{fig:hole}. The positional encoding method successfully fills these holes, accurately replicating the correct shape. This effectiveness is attributed to its inherent periodic property, which provides prior information. However, a limitation of this approach is the lack of smoothness on the road surface. Multi-resolution hash positional encoding excels in rendering the road surface with remarkable smoothness. It fills the horizontal holes smoothly and attempts to fill the corner holes with values from adjacent areas. This is due to the absence of prior information in the multi-resolution hash positional encoding approach, leading to a less smooth appearance. We also examine the ability of multi-resolution hash positional encoding when it has only a single resolution or devoid hash function. The figure shows that only a single hash function without multiresolution cannot fill horizontal and corner holes. Conversely, employing a multiresolution approach without a hash function demonstrates a filling capability similar to the multi-resolution hash positional encoding method.

\subsection{Color and Semantic Reconstruction with Different PE methods and Ground Truth Height }
\begin{figure}[tb]
  \centering
  \raisebox{0.1in}{\rotatebox[origin=lc]{90}{\tiny SfM sparse \hspace{6.5em} \tiny SfM dense \hspace{6.5em} \tiny LiDAR \hspace{6.5em} \tiny Vehicle camera pose}}
  \includegraphics[scale=0.25]{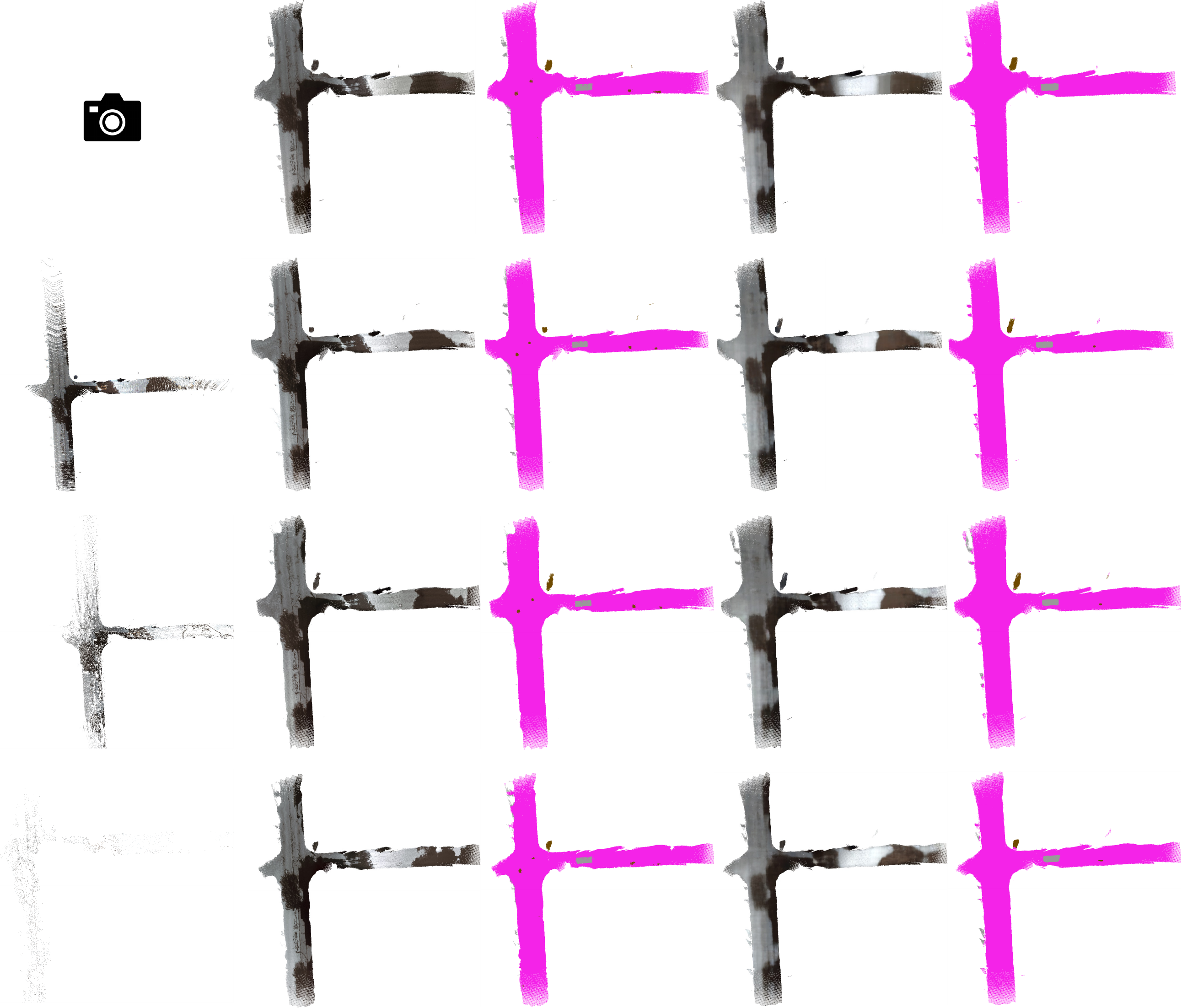}
  \\
  \hspace{1em} \tiny Input \hspace{4em} \tiny Hash PE color \hspace{2em} \tiny Hash PE semantic \hspace{2em} \tiny PE color \hspace{4em} \tiny PE semantic
  \vspace*{2\baselineskip}
  \\
  \includegraphics[scale=0.25]{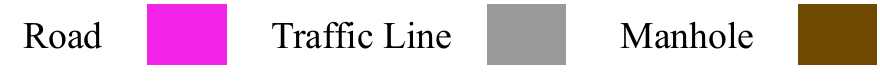}
  \caption{In qualitative comparison with the PE and Hash PE in the datasets of the vehicle camera pose, LiDAR points, SfMs dense points, and SfMs sparse points. The first column shows each input dataset, and the rest illustrates the reconstruction result. Hash PE has a better render quality in each result than in color and semantics.
  }
  \label{fig:render_r}
\end{figure}

In this part, as shown in \cref{fig:render_r}, we show our experiment results in color and semantic results in different positional encoding methods for our method \textbf{NeRO} with different input datasets, and in \cref{tab:expre}, we offer our quantitive metric for each result. 

The results show that road surface height reconstruction with lidar point clouds performs best over other road height sources. However, the road surface obtained by SFM reconstruction is similar to that of vehicle pose, probably caused by the fact that the Kitti dataset is collected on flat roads. We also show that the performs of hash PE is better than PE.

\begin{table}[tb]
  \caption{Quantitive result for our method in reconstructing the road surface in color and semantics with the road height from vehicle camera pose, LiDAR, and SfMs using multi-resolution hash positional encoding or positional encoding method.
  }
  \label{tab:expre}
  \centering
  \begin{tabular}{@{}llll@{}}
    \toprule
    Road height & Positional Encoding Type & PSNR & mIoU\\
    \midrule
    Vehicle Camera Pose  & PE & 17.81 & 0.704 \\
    Vehicle Camera Pose  & Hash PE & 25.73 & 0.988\\
    LiDAR & PE & 18.87 & 0.701 \\
    LiDAR & Hash PE & 29.20 & 0.994 \\
    SfM-Dense & PE & 18.76 & 0.784 \\
    SfM-Dense & Hash PE & 25.81 & 0.975 \\
    SfM-Sparse & PE & 18.38 & 0.780 \\
    SfM-Sparse & Hash PE & 24.37 & 0.967 \\
  \bottomrule
  \end{tabular}
\end{table}

% 下面三个部份合一，变成一个color语义效果在不同pe下

\subsection{Road Semantic Reconstruction with Noise Labels}
This section shows how our method is applied in some applications, like sparse semantic labels and semantic label denoising. In NeRF-based models, like \cite{wang2023semantic,zhi2021place} offer its ability on that kind of application. We only chose the LiDAR road height for this process because it has relatively accurate road height values, and in \cref{tab:semsp} and \cref{tab:semde}, we offer our quantitive metric for each result. 

\subsubsection{Sparse Label}
\begin{figure}[tb]
  \centering
  \raisebox{0.2in}{\rotatebox[origin=lc]{90}{\tiny 10\% semantic labels}}
  \includegraphics[scale=0.4]{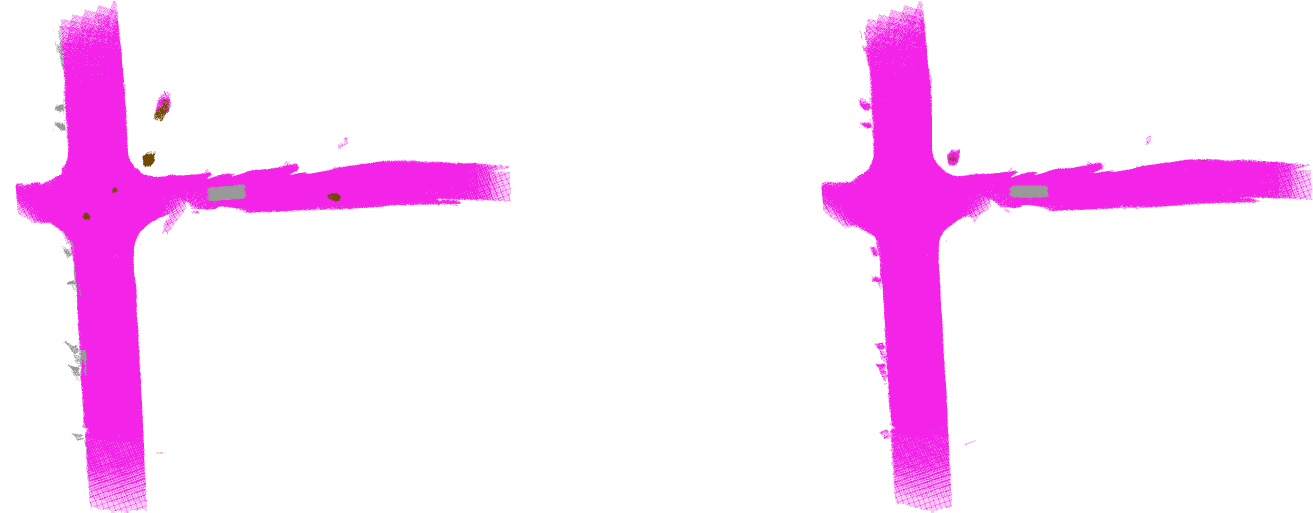}
  \\
  \hspace{-1.8em} \tiny Hash PE semantic \hspace{14em} \tiny PE semantic
  \vspace*{2\baselineskip}
  \\
  \includegraphics[scale=0.25]{sem_box.pdf}
  \caption{Comparison with the positional encoding and multi-resolution hash positional encoding in the sparse semantic labels. Both methods render the whole structure of the road, but the Hash PE on the left gives more detail in the result.
  }
  \label{fig:sparse_r}
\end{figure}

We hypothesize a scenario where a significant portion of ground truth semantic images is unavailable, leaving only a minimal subset for use. From \cite{zhi2021place}, it indicated that using less than 10\% of semantic images will significantly decrease the rendering quality. Our investigation will set the threshold to only 10\% of images to evaluate whether our method could reconstruct the road surface. \cref{fig:sparse_r} presents the outcomes by using positional encoding or multi-resolution hash positional encoding. The results demonstrate that, despite the limited availability of training semantics, both methods can reconstruct the road surfaces. However, the distinct advantage of multi-resolution hash positional encoding is its ability to render detailed features, such as manholes, which positional encoding fails to replicate with the same level of detail. \cref{tab:semsp} shows the quantitive result in the mIoU. Here, it is evident that multi-resolution hash positional encoding outperforms positional encoding, showcasing superior mIoU values. 

\begin{table}[tb]
  \caption{Quantitive result for our method for sparse labels, when using only 10\% training semantic dataset.
  }
  \label{tab:semsp}
  \centering
  \begin{tabular}{@{}lll@{}}
    \toprule
    Sparse Ratio & Positional Encoding Type & mIoU\\
    \midrule
    0.1  & PE & 0.670 \\
    0.1  & Hash PE & 0.823\\
  \bottomrule
  \end{tabular}
\end{table}

\subsubsection{Noise Label}
\begin{figure}[tb]
  \centering
  \raisebox{0.3in}{\rotatebox[origin=lc]{90}{\tiny 50\% denoised labels
 \hspace{8.8em} \tiny 90\% denoised labels}}
  \includegraphics[scale=0.4]{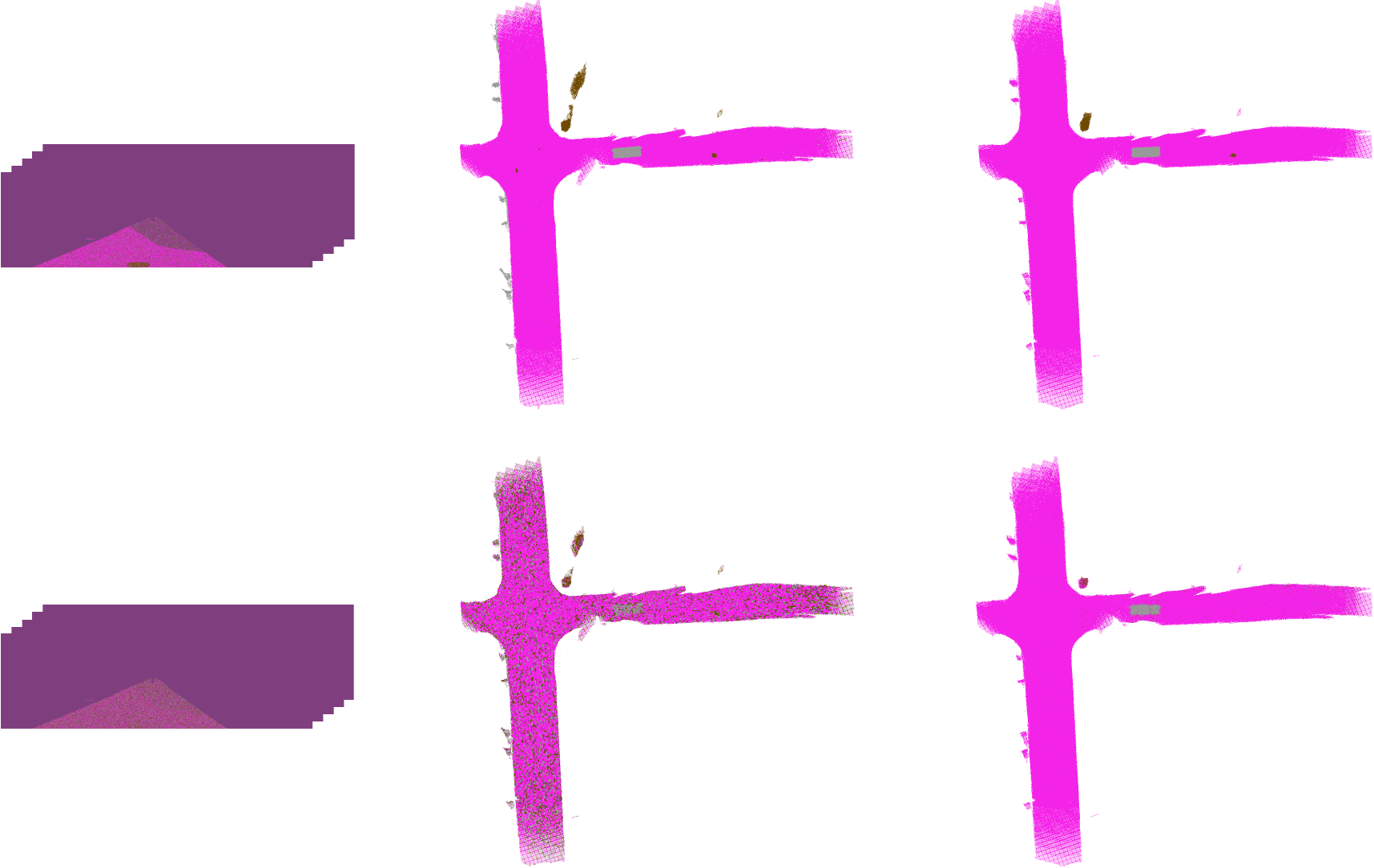}
  \\
  \tiny Input \hspace{12em} \tiny Hash PE semantic \hspace{10em} \tiny PE semantic
  \vspace*{2\baselineskip}
  \\
  \includegraphics[scale=0.25]{sem_box.pdf}
  \caption{Comparison with the positional encoding and multi-resolution hash positional encoding in the noisy semantic labels. The first column shows the input noisy datasets, the second column records the multi-resolution hash positional encoding result, and the final column gives the positional encoding result. 
  }
  \label{fig:denoise_r}
\end{figure}

In the actual scenarios, the road datasets have noise content, so we simulate some pixel noises to the ground truth by using the method in \cite{zhi2021place}, and the noised semantic labels are used to examine our method's denoise ability. \cref{fig:denoise_r} illustrates the outcomes of applying denoising reconstruction to noisy labels using positional encoding or multi-resolution hash positional encoding.  In scenarios with 50\% noise in the labels, it is observed that both multi-resolution hash positional encoding and positional encoding can reconstruct the road surface. Still, the positional encoding method fails to reconstruct some specific items, like manholes in the middle of the road. When we add the ratio of noise labels to 90\%, multi-resolution hash positional encoding cannot denoise the label due to its accurate learning ability. Its precision in learning leads to the unintentional incorporation of noise labels as road surface components. In contrast, the positional encoding method continues to denoise the labels. However, it can still not render detailed features like manholes in the middle of the road. The quantitative analysis in \cref{tab:semde} shows that the mIoU value of positional encoding is higher than multi-resolution hash positional encoding, indicating a better semantic denoise ability. Still, positional encoding needs to be more comprehensive in rendering the details. Further investigations were conducted to determine the noise ratio threshold of the denoise ability of multi-resolution hash positional encoding. Visual results indicate that multi-resolution hash positional encoding begins to integrate noise labels into the road surface when the noise ratio exceeds 0.6.

\begin{table}[tb]
  \caption{The quantitive result for our method is denoising the noisy semantic labels, which use positional encoding or multi-resolution hash positional encoding method to denoise the label with the noise ratio when 50\% or 90\% labels are noised in LiDAR datasets.
  }
  \label{tab:semde}
  \centering
  \begin{tabular}{@{}lll@{}}
    \toprule
    Noise Ratio & Positional Encoding Type & mIoU\\
    \midrule
    0.5  & PE & 0.693 \\
    0.5  & Hash PE & 0.625\\
    0.9  & PE & 0.420 \\
    0.9  & Hash PE & 0.292\\
  \bottomrule
  \end{tabular}
\end{table}

\subsection{Training speed}
\begin{figure}[tb]
  \centering
  \includegraphics[scale=0.28]{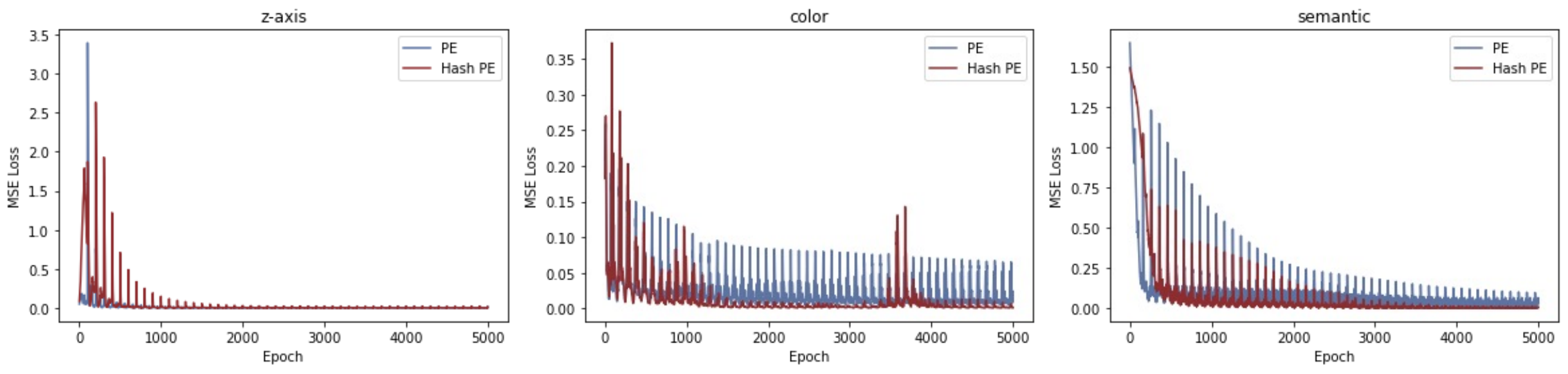}
  \caption{Comparison in the training loss decay between positional encoding and multi-resolution hash positional encoding in the z-axis, color, and semantic. Those spikes appear because we train in sequence, not in random.
  }
  \label{fig:loss_r}
\end{figure}

During the networks' training phase, the loss reduction is shown in the \cref{fig:loss_r}. When examining the vertical z-axis loss, both positional encoding and multi-resolution hash positional encoding methods exhibit comparable training speeds, ultimately converging on their global minima. Multi-resolution hash positional encoding demonstrates an accelerated color and semantic loss training speed, efficiently attaining the global minima. Conversely, the positional encoding method experiences a slower decrease in loss, facing challenges in achieving convergence to the global minima.

\section{Limitation}
The limitation of our method is the difficulty in achieving a balance between the ability to fit detailed information and model complexity on road surfaces height reconstruction. Most road surface is flat and could be described with low parameter multivariate equations, and there is no need to use MLPs with big parameters. But for the sake of some little detailed information like holes in the road we have to use position encoding and MLPs with huge parameters. In future research, we propose to refine our approach by a hybrid representation of road surface.

\section{Conclusion}
In conclusion, we introduce an position encoding MLPs-based neural road reconstruction method that accepts the various sources of round truth of height to render the color and semantic information with road height output. Our experiment shows the success of our rendering ability in either color or semantic information. In the semantic applications, it also shows that it can handle sparse labels and noise labels. We also compare the Positional Encoding and Muti-resolution Positional Encoding methods to deliver each performance. This indicates that the Muti-resolution Positional Encoding method performs better in rendering the road surfaces in quality and speed.

\newpage
\bibliographystyle{splncs04}
\bibliography{egbib}
\end{document}